\documentclass[]{interact}

\usepackage{epstopdf}
\usepackage[caption=false]{subfig}
\usepackage{xurl}

\usepackage{array}
\usepackage[flushleft]{threeparttable} 

\newcolumntype{P}[1]{>{\centering\arraybackslash}p{#1}}

\usepackage[numbers]{natbib}

\theoremstyle{plain}

\theoremstyle{definition}

\theoremstyle{remark}

\begin{document}

\articletype{Original Research Paper}

\title{Performance Insights-based AI-driven Football Transfer Fee Prediction}

\author{
\name{D. Sulimov\textsuperscript{a}\thanks{CONTACT D. Sulimov. Email: dasulimov@edu.hse.ru}}
\affil{\textsuperscript{a}HSE University,
Moscow, Russia}
}

\maketitle

\begin{abstract}
We developed an artificial intelligence approach to predict the transfer fee of a football player. This model can help clubs make better decisions about which players to buy and sell, which can lead to improved performance and increased club budgets.
We collected data on player performance, transfer fees, and other factors that might affect a player's value. We then used this data to train a machine learning model that can accurately predict a player's impact on the game. We further passed the obtained results as one of the features to the predictor of transfer fees.
The model can help clubs identify players who are undervalued and who could be sold for a profit. It can also help clubs avoid overpaying for players who are not worth the money.
We believe that our model can be a valuable tool for football clubs. It can help them make better decisions about player recruitment and transfers.
\end{abstract}


\section{Introduction} \label{intro}
In the context of football evolving into a lucrative industry, players could be regarded as the most valuable assets of a club. The aforementioned study\citep{Miao} has effectively established the statistical significance of the relationship between a football player’s performance and market value using the Lasso Regression methodology. Consequently, armed with insights into a player's performance, clubs can strategically engage in advantageous transactions, augmenting their revenue streams.

Recently, various approaches have emerged for estimating a player’s transfer value using the PA framework. These approaches encompass a range of methodologies, including regression analysis methods\citep{Poli2022, Ian2023}, where fundamental actions like passes and goals are considered. Additional techniques such as machine learning\citep{Ian2023, Roland2022} have been employed, with authors exploring factors like the \emph{plus-minus} rating of footballers. Deep learning approaches\citep{Patnaik2019} have also been applied, emphasising relative and absolute PA parameters, such as successful tackles, dribbles, and more. Notably, this research also incorporates media activity as a factor within the transfer value prediction model, including metrics like the number of \emph{YouTube} videos, \emph{Reddit} posts, and other indicators of player popularity.

In general, there could be distinguished a bunch of factors, having an impact on the transfer values. We divide them into two groups: major and minor.

Regarding minors, we can name media activity, relationships with coaches, out-of-pitch behaviour, etc. Technically, these are not quantitative factors that could not be calculated directly. While the research\citep{Patnaik2019} presents an intriguing method for estimating media activity by considering the total views of YouTube videos, leveraging Google trends data, etc., related to a player, there may be more accurate approaches. Indeed, these metrics may only partially capture the actual number of people genuinely engaged in a footballer’s activity. The reason behind this limitation lies in the fact that these metrics primarily measure online interactions and views, which may not necessarily represent the complete scope of a player’s popularity and impact on their fans or the football community as a whole.

For the former, we can list age, the remaining term of a contract, pitch role, etc. However, the most significant parameter among major factors is a player’s contribution to the team. Indeed, correct assessment of player’s actions is relevant not only in the case of transfer fee prediction but also is highly useful for both scouts to recruit players and coaches to form a team roster.\par

Also PA could be used as a tool of significant feedback\citep{Miller2012,Lasse2022,Nicholls2018}.In order to evaluate performance scores, various models have been presented up-to-date. Over the past years, deep learning methods of players' actions analysis have gained popularity\citep{Rahimian2022,Guiliang2018,Javier2019}. However, they mostly take into account actions with direct influence on the game (e.g., key passes, goals, etc.), yet they may need to reflect the player's essence accurately. Such approaches are considered in \citep{Anzer2021,Power2017,REIN2017,Smit2018,Lucey2015QualityVQ}. For instance, some actions do not lead to scoring a goal themselves. However, they contribute a lot to perspective scoring (e.g. starting a pass of an attack, which opens some space for the offensive purpose, successful dribbling, etc.). Techniques for evaluating this staff are described in \citep{VAN, Liu2020}, opening up a view of the probabilistic approach to footballer's performance score evaluation and finding its applications in detecting young talents and substitutions of players left a club.\par
The study builds upon the Bayesian approach, as established in work \citep{VAN}, yet reconstructs the modelling approach and expands the application part. This research focuses on utilising event-stream data and incorporates a long-run perspective when assessing the impact of a player’s attacking actions within a possession chain that culminates in a shot on goal. The possession chain itself will be elaborated upon in the subsequent section. Furthermore, the study introduces a unified algorithm that enables the prediction of transfer fee value gain or loss within a specified time. This prediction is facilitated by developing supplementary models specifically designed for computing the performance score. Therefore, by leveraging the Bayesian approach and employing event-stream data, this study provides a framework for evaluating player performance within possession chains and offers a predictive algorithm for assessing a player’s potential transfer fee value fluctuation. This serves as a valuable resource for coaches and club managers and contributes to the broader understanding of player evaluation and decision-making processes in the world of football.

To evaluate a player’s performance, the researchers analyse a sample of possession chains in which the player has participated, ultimately computing their performance score. This score serves as a measure of the player’s effectiveness in contributing to goal-scoring opportunities.

\section{Materials and methods}
\label{method}
Since we do not provide open access to the developed programming code, we do not conduct comparison with related works.

\begin{figure}[h!]
    \centering
    \includegraphics[scale=0.5]{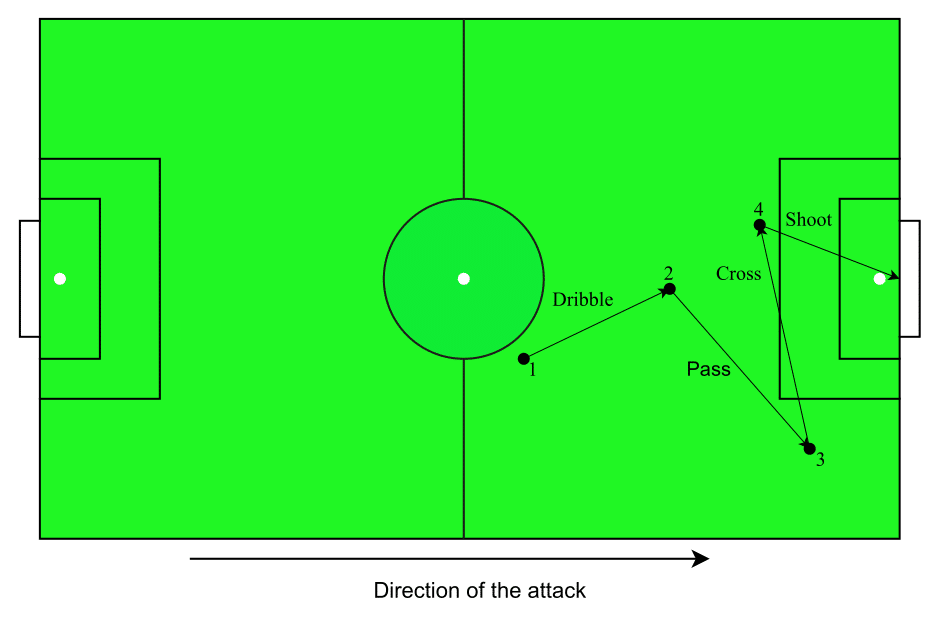}
    \caption{\centering{Possession chain.Black points represent ball's position,\par arrows - direction of the actions performed over it,\par signed next to each arrow.}}
    \label{fig:pos_chain}
\end{figure}

The ultimate target of every team during a game is to end a match by scoring as many goals as possible. Let us consider an attack by a random team. This attack could be represented as a possession chain. We note such a chain as a sequence of consecutive actions of a given team, possessing a ball without any interception from the opponents and finishing with a shot to the rival’s goals. The example of a possession chain is illustrated in Figure \ref{fig:pos_chain}. The whole chain could also be split into separate actions with the total number of $K$.
Throughout a possession chain. We will focus on the ball movement: the ball changes its state with every action performed on it. The state can be represented as a vector \begin{equation}\label{eq:fourth}{\Vec{s}_{k}} = (x_{k}, y_{k})^\intercal\end{equation}where $x_{k}$ and $y_{k}$ are the coordinates of a current ball's position on the pitch, and $k$ is the index of an action done over the ball. Every ball’s position $\Vec{s}_{k}$ on the pitch over a given chain $j$ implies different probability to score. Let \begin{equation}\label{eq:third}\mathbf{P}(score|\Vec{s}_{k})\end{equation} be the probability of scoring a goal, given the ball, being in the state $\Vec{s}_{k}$. Then, a footballer $i$, performing an action $k$ over the ball during the chain $j$, moving it from state $\Vec{s}_{k}$ to $\Vec{s}_{k+1}$ receives a score \begin{equation}\label{eq:fifth}c_{i}(k)(j) = \mathbf{P}(score|\Vec{s}_{k+1}) - \mathbf{P}(score|\Vec{s}_{k})\end{equation}. There could be three possible outcomes:\par
1) $c_{i}(k)(j)>0$: the player $i$ receives a reward with a positive impact;\par
2) $c_{i}(k)(j)<0$: the player $i$ receives a penalty with a negative impact;\par
3) $c_{i}(k)(j)=0$: the player $i$ receives neither a reward or penalty with a neutral impact.\par
Hence, a total contribution of the player $i$ within a possession chain $j$ could be formulated as: \begin{equation}\label{eq:sixth}C_{i}(j)=\Sigma_{k=1}^{K_{j,i}}c_{i}(k)(j)\end{equation}where $K_{j,i}$ represents the total number of actions, completed by the player $i$ within the one possession chain $j$. Thus, we can derive a total score, assigned to a player for one match: \begin{equation}\label{eq:seventh}C_{i}=\Sigma_{j=1}^{J_{i}}\Sigma_{k=1}^{K_{j,i}}c_{i}(k)(j)\end{equation}where $J_{i}$ is the total number of possession chains, in which the player $i$ has played a part.\par
Since each footballer plays a distinct number of matches, for further comparison of players between each other, the normalising correction is applied:
\begin{equation}
    \label{eq:normalising}C_{i,norm}=\frac{1}{N_{i}}*\Sigma_{j=1}^{J_{i}}\Sigma_{k=1}^{K_{j,i}}c_{i}(k)(j)
\end{equation}\par
where $N_{i}$ is the number of games, in which the player $i$ has taken part.\par
As a reward for the player, who performs the last action of the possession chain, a self-developed expected goals model (xG) was applied. With the same idea, being a probabilistic classifier, the model gives a probability to score with the last action, $K_{j}$, i.e. a shoot, in the chain $j$:\begin{equation}\label{eg:eighth}c_{xG}=\mathbf{P}(score|action=K_{j,i})\end{equation}In the case of performing the very last action $K_{j,i}$, a footballer gets either $c_{xG}$ as a penalty or (1-$c_{xG}$) as a reward.\par
Also, the model considers players' pitch roles (e.g. a defender, a midfielder or a striker). This implies mapping a player to his position, according to official formation. The sizes of a pitch are considered to be 120m x 80m and it is divided into 3 zones: defending, midfield and attacking. Each zone is separated by 40 meters in the direction of an attack, presented in Figure \ref{fig:pos_chain}. The weights to a footballer's score are set the following way:\par
1) A defender has no additional reward for actions in defending zone, 1.5x for successful actions in midfield zone and 2x for activity in the attacking part of the pitch;\par
2) A midfielder gets 1.5x for successful participation in a possession chain in the attacking zone;\par
3) A striker receives no reward for any actions, regardless of the zone.\par
Hence, to emphasise a footballer's ability to contribute to attacking actions, his score is multiplied by the weight per the player's pitch position. This advanced feature mainly serves as an assistant in forming the attacking line of a team.\par
\par
The data that support the findings of this study are available in open-data at \url{https://github.com/statsbomb/open-data}, reference \cite{statsbomb}, provided by \emph{StatsBomb}, a sports analytic company. The provided data set consists of more than one thousand games, encompassing national and international competitions for men’s and women’s football. Each match’s data consists of 105 features, describing every event of the game in the context of timestamp, type of the action, duration of the action, etc.

\subsection{xG model development}
\label{xG}
The objective of this section is to develop an xG (expected goals) model based on the event-stream data. In essence, the xG score is the probability of scoring with a shoot being under specific circumstances. These could be several factors (e.g. distance to the goals, shot technique, etc.), presented in the Table \ref{tab:xg_features}.
\begin{table}[h!]
    \tbl{Features for xG model.}
    {\begin{tabular}{|P{6.7cm}|P{6.7cm}|}
    \hline
    \textbf{Categorical features} & \textbf{Continuous features}\\
    \hline
    Shot technique & Shooting angle\\
    \hline
    Shot body part & Distance to the goals\\
    \hline
    Shot type & \\
    \hline
    Whether a player is under the pressure of an opponent or not & \\
    \hline
    \end{tabular}}
    \label{tab:xg_features}
\end{table}\par

Let us define features "shooting angle" and "distance to the goals" explicitly:\par
1) Shooting angle ($\theta$) - angle, which is formed by the player's view to the goals between two goalposts. There are three cases of possible position of a player and each of them preserves specific calculations:\par
\begin{figure}[h!]
\centering
\subfloat[$y_{k}>40$. Red dot represents the centre of the goals.]{%
\resizebox*{8.0cm}{!}{\includegraphics{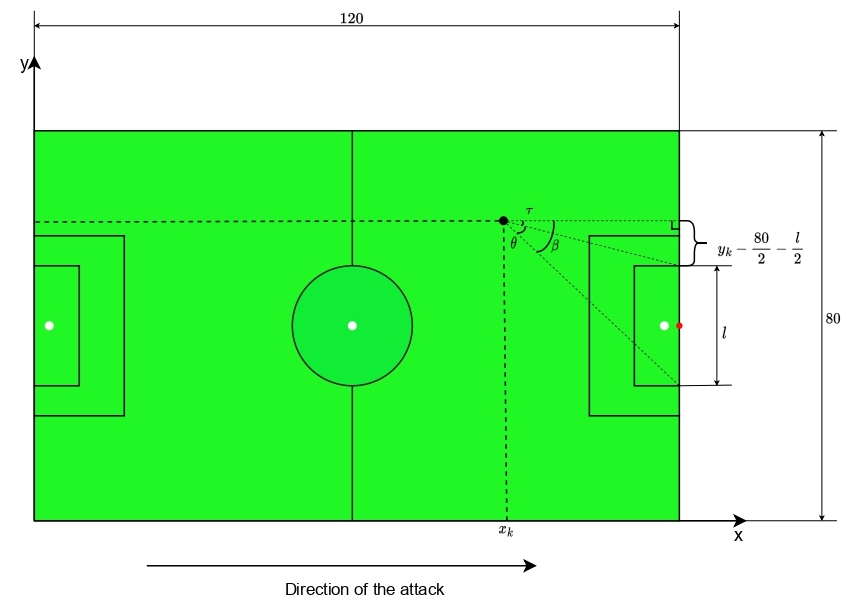}}}\hspace{5pt}
\subfloat[$y_{k}<40$.]{%
\resizebox*{8.0cm}{!}{\includegraphics{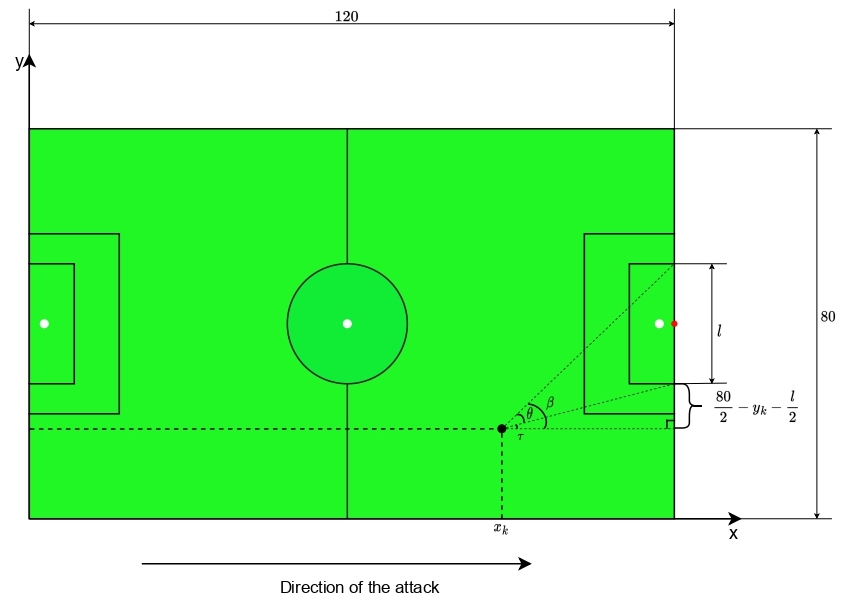}}}
\subfloat[$y_{k}=40$.]{%
\resizebox*{8.0cm}{!}{\includegraphics{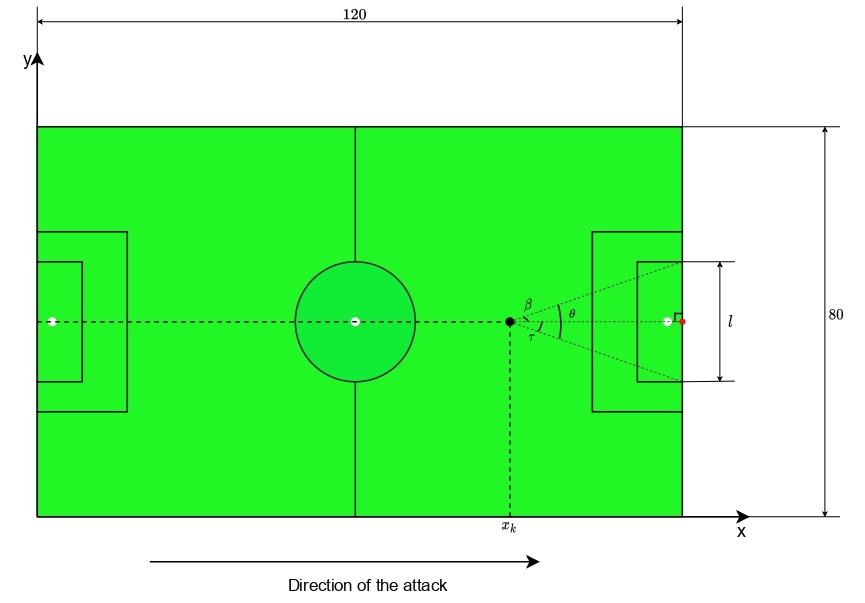}}}
\caption{Examples of a shooting angle from different views.} \label{equals40}
\end{figure}

\begin{itemize}
    \item $y_{k} > 40$ (presented in Figure \ref{equals40}a): $\theta = \arctan({\beta}) - \arctan(\tau) = \arctan(\frac{y_{k} - \frac{80}{2} + \frac{l}{2}}{120-x_{k}}) - \arctan(\frac{y_{k} - \frac{80}{2} - \frac{l}{2}}{120-x_{k}})$
    \item $y_{k} < 40$ (presented in Figure \ref{equals40}b): $\theta = \arctan({\beta}) - \arctan(\tau) = \arctan(\frac{\frac{80}{2} + \frac{l}{2} - y_{k}}{120-x_{k}}) - \arctan(\frac{\frac{80}{2} - \frac{l}{2} - y_{k}}{120-x_{k}})$
    \item $y_{k} = 40$ (presented in Figure \ref{equals40}c): $\theta = \arctan(\beta) + \arctan(\tau) = \arctan(\frac{\frac{l}{2}}{120-x_{k}}) + \arctan(\frac{\frac{l}{2}}{120-x_{k}}) = 2\arctan(\frac{\frac{l}{2}}{120-x_{k}})$
\end{itemize}\newpage

\begin{figure}[h!]
    \centering
    \includegraphics[scale=0.55]{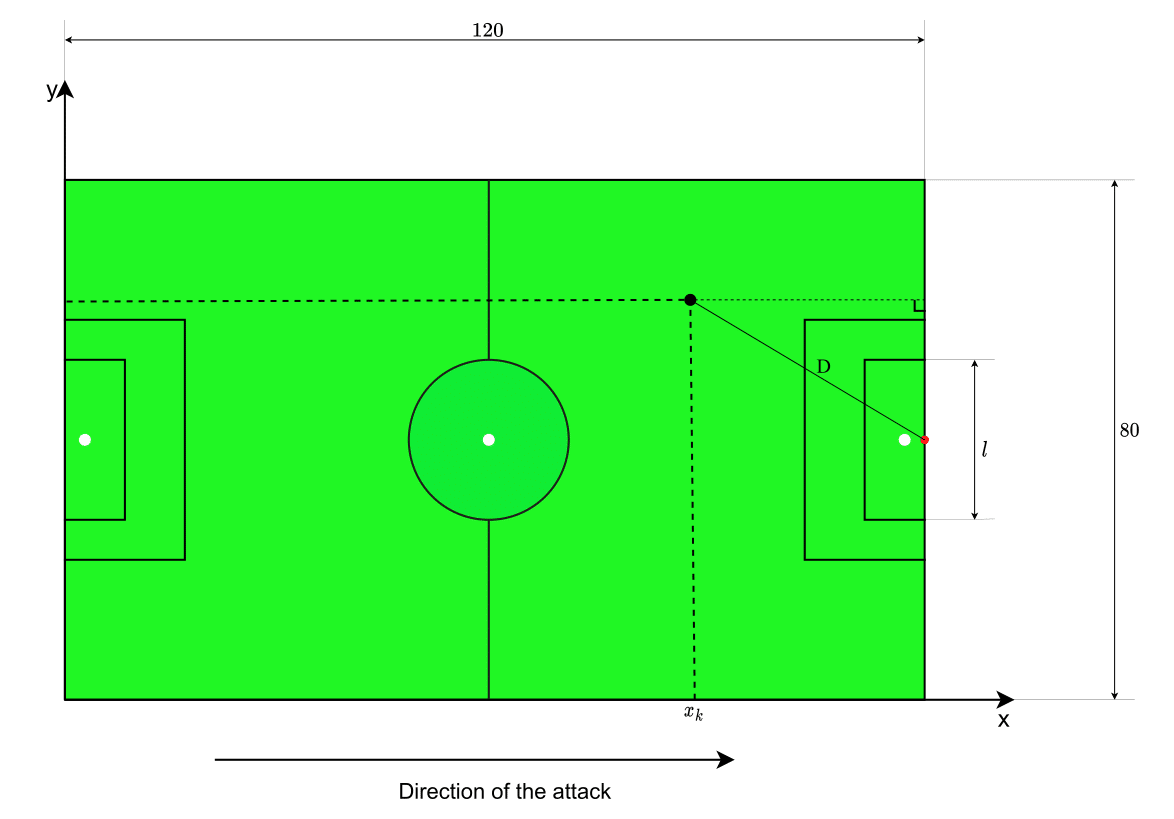}
    \caption{Distance to the goals. Red dot represents the centre of the goals.}
    \label{fig:distance}
\end{figure}

2) Distance to the goals ($D$) - distance from a player to the centre of the opponent's goals, presented in Figure \ref{fig:distance}. It is a spatial separation between the player and the target area: $D = \sqrt{{x_{k}}^2 + ({\frac{80}{2}-y_{k}})^2}$. There is a base case when $y_{k}=40$: $D = 120-x_{k}$.\par

The task at hand pertains to a binary classification problem with imbalanced classes. Specifically, the labels are assigned to shots, categorised as 1 if they result in a score and 0 otherwise. The overall proportion of class 1 instances in the provided data set is 0.22, warranting the assignment of a corresponding weight to class 0. To develop the xG model, the Python language is selected, leveraging the scikit-learn package and its built-in implementations of algorithms such as \emph{Logistic Regression}, \emph{Random Forest Classifier}, and \emph{CatBoost Classifier}. 

\emph{Logistic Regression} is often employed as a baseline solution for binary classification problems due to its simplicity and interpretability. It is commonly utilised to establish a performance benchmark against more complex algorithms. 

On the other hand, \emph{Random Forest Classifier} is known for its ability to handle high-dimensional data without requiring dimensionality reduction techniques.

In the xG model, \emph{CatBoost Classifier} is a robust algorithm that stands out due to its robustness in handling missing data and its native ability to handle categorical features. This feature makes \emph{CatBoost} particularly beneficial when dealing with data sets that contain a substantial number of categorical variables, as is often the case in the xG model.

Given the presence of imbalanced data, the evaluation metrics employed include the weighted variants of \emph{F1-score}, \emph{precision}, \emph{recall}, and the \emph{AUC} score. The data is partitioned into two groups for model training and testing, with the selected proportions of 0.7 and 0.3, respectively. Stratification is employed based on the target variable to ensure the preservation of class distribution during the partitioning process. Furthermore, grid search is chosen as the method for hyperparameter tuning. The results of the models, trained on the algorithms, are presented in Table \ref{tab:xG}.
\begin{table}[h!]
    \tbl{xG models' metrics.}
    {\begin{tabular}{|P{2cm}|P{3cm}|P{3cm}|P{2cm}|P{2cm}|}
    \hline
     & F1-score(weighted) & Precision(weighted) & Recall(weighted) & AUC score\\
    \hline
    Logistic Regression & 0.74 & 0.88 & 0.69 & \textbf{0.85}\\
    \hline
    Random Forest Classifier & \textbf{0.86} & 0.85 & 0.86 & 0.73\\
    \hline
    CatBoost Classifier & \textbf{0.86} & \textbf{0.89} & \textbf{0.88} & 0.78\\
    \hline
    \end{tabular}}
    \label{tab:xG}
\end{table}\par
Our investigation's primary focus centres around identifying all positive instances, thus prioritising the \emph{recall} metric. Accordingly, the \emph{CatBoost Classifier} demonstrates its superiority over alternative algorithms. As part of the analysis, we examined the \emph{precision-recall} curve to understand how the model's \emph{precision} and \emph{recall} values change across various decision thresholds.

For instance, the \emph{precision-recall} curve of the \emph{CatBoost Classifier} algorithm, as depicted in Figure \ref{pr_curves}a, exhibits an initial sharp increase followed by a gradual decrease. This observation suggests a high level of \emph{precision}, albeit at the cost of a slightly reduced \emph{recall}. These findings are consistent with the results obtained during model training.

\subsection{Goal-scoring predictor development} 
\label{scoring_predictor}
The objective of the goal-scoring predictor is to estimate the probability outlined in Section \ref{method}, precisely (\ref{eq:fifth}). Due to our setting, the predictor plays
an essential role in the whole study. Therefore, its characteristics must meet high evaluation standards to accurately assess the score assigned to the player with the most precise prediction. Each action of a possession chain, excluding the last one, described in \ref{xG}, is viewed through the following factors in Table \ref{tab:scoring_features}.
\begin{table}[h!]
    \tbl{Features for goal-scoring predictor.}
    {\begin{tabular}{|P{4.3cm}|P{4.3cm}|P{4.3cm}|}
    \hline
    \textbf{Categorical features} & \textbf{Discrete features} & \textbf{Continuous features}\\
    \hline
    Type of the action & Coordinates of the action & Duration of the action\\
    \hline
    Whether the action is performed under the pressure of an opponent & Number of actions from the current position to the end of the possession chain & Cumulative duration of the chain before the action\\
    \hline
    Type of the previous action & Timestamp of the action &\\
    \hline
    Play pattern* & Number of the action in the chain &\\
    \hline
    \end{tabular}}
    \label{tab:scoring_features}
    \begin{tablenotes}
        \small
        \item * - name of the play pattern relevant to the action, e.g. "from corner": the action is a part of the passage of play following a corner, "from goal kick": the action is a part of the passage of play following a keeper distribution, etc.
    \end{tablenotes}
\end{table}\par
The outcome of each action under consideration falls into one of two categories: scoring (=1) or not scoring (=0). Thus, we are confronted with a binary classification problem. However, we encounter the challenge of highly imbalanced classes once again, as discussed in Section \ref{xG}. Since then, the algorithms are used. Among all the actions included in the data set, only 12\% of them ultimately lead to scoring. Consequently, class 0 is assigned a weight of 0.12 to address the class imbalance issue. The results are presented in Table \ref{tab:scoring_predictor_table}.
\begin{table}[h!]
    \tbl{Goal-scoring predictor models' metrics.}
        {\begin{tabular}{|P{2cm}|P{3cm}|P{3cm}|P{3cm}|P{1cm}|}
    \hline
     & F1-score(weighted) & Precision(weighted) & Recall(weighted) & AUC score\\
    \hline
    Logistic Regression & 0.44 & 0.83 & 0.37 & 0.51\\
    \hline
    Random Forest Classifier & 0.66 & 0.88 & 0.59 & 0.79\\
    \hline
    CatBoost Classifier & \textbf{0.95} & \textbf{0.95} & \textbf{0.95} & \textbf{0.97}\\
    \hline
    \end{tabular}}
    \label{tab:scoring_predictor_table}
\end{table}\par
\begin{figure}[h!]
\centering
\subfloat[Precision-recall curves of xG models.]{%
\resizebox*{7cm}{!}{\includegraphics{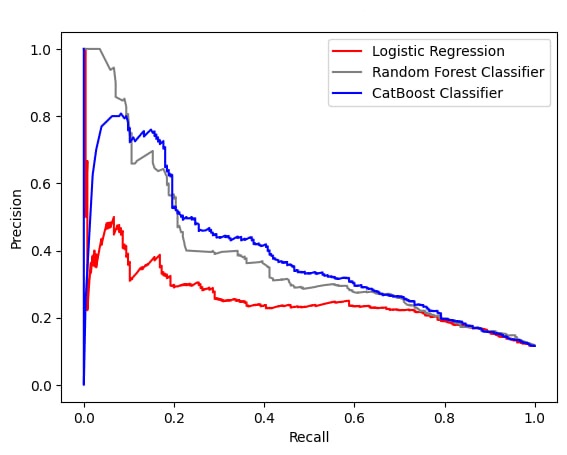}}}\hspace{5pt}
\subfloat[Precision-recall curves of goal-scoring predictor models.]{%
\resizebox*{7cm}{!}{\includegraphics{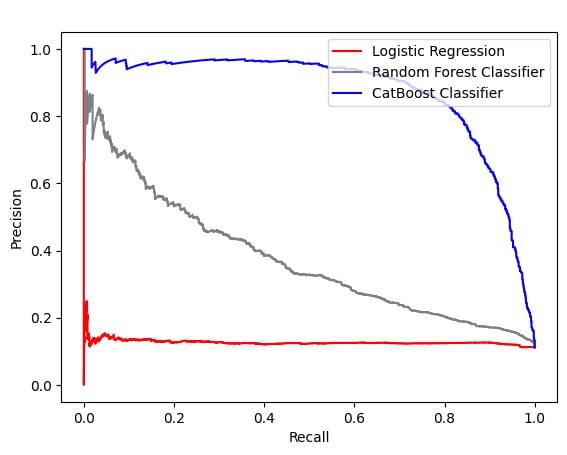}}}
\caption{Precision-recall curves.} \label{pr_curves}
\end{figure}

Similarly to the xG model, the \emph{CatBoost Classifier} is the most effective among all the algorithms considered, consistently achieving top performance across all evaluation metrics. The \emph{precision-recall} curve, as presented in Figure \ref{pr_curves}b, illustrates that the \emph{CatBoost Classifier} exhibits the most significant area under the curve compared to the other two algorithms. This indicates the algorithm's ability to detect instances belonging to the minority class. The \emph{precision-recall} curve visualisation serves as compelling evidence of the model’s high accuracy in identifying relevant items.

By leveraging the \emph{CatBoost Classifier} algorithm, the goal-scoring predictor achieves a remarkable performance, exceeding its counterparts in terms of \emph{precision}, \emph{recall}, and overall model capability. The visualisation of the \emph{precision-recall} curve strongly supports these findings, highlighting the beneficial qualities of the algorithm in accurately identifying relevant instances. The visualisation is presented in Figure \ref{pr_curves}.\newpage

\subsection{Development of forecasting model for transfer fees} \label{Forecasting}
As discussed in Section \ref{intro}, several factors affect footballers' transfer values. Our target variable is the gain or loss in transfer value over a certain period. We are particularly interested in the relationship between this variable and the player score computed in Subsection \ref{scoring_predictor}. Nevertheless, the specification for the model training includes other features, the whole list is given in Table \ref{tab:transfer_value_features}.
\begin{table}[h!]
    \tbl{Features for transfer value gain/loss predictor.}
    {\begin{tabular}{|P{6.7cm}|P{6.7cm}|}
    \hline
    \textbf{Categorical features} & \textbf{Discrete features}\\
    \hline
    Pitch position of a player & Evaluation time of a player(number of games)\\
    \hline
     & Time lag between transfer value change (months)\\
    \hline
     & Age of a player\\
    \hline
     & Age$^2$ of a player\\
    \hline
    \end{tabular}}
    \label{tab:transfer_value_features}
\end{table}\par

The task revolves around predicting the increase or decrease in transfer fees, measured in millions of euros, constituting a regression problem. To address this, we employ regression algorithms, precisely the \emph{Random Forest Regressor}, \emph{CatBoost Regressor}, and \emph{Decision Tree Regressor}. 

All the algorithms above exhibit robustness in the presence of outliers. Furthermore, the \emph{Decision Tree Regressor}, along with the \emph{Random Forest Regressor} and \emph{CatBoost Regressor}, possess the capacity to capture nonlinear relationships and interactions among features in the data set. Moreover, the \emph{CatBoost Regressor} includes an automated feature scaling mechanism, thereby obviating the necessity for manual feature scaling or normalisation. This capability streamlines the preprocessing stage and alleviates the associated time and effort requirements.

The evaluation of our research is based on appropriate metrics for regression tasks, namely the \emph{root mean squared error}(RMSE) and \emph{mean absolute error}(MAE). These metrics allow us to assess the accuracy and precision of our models in predicting the gain or loss in transfer fees.

Furthermore, we utilise each algorithm's built-in \emph{feature\_importance} method to determine the relative importance of features in relation to the player’s score and the gain or loss in transfer value. This analysis will allow us to rank the level of importance among all the features.

Similarly to the approach employed in Sections \ref{xG} and \ref{scoring_predictor}, the data set is split into training and testing sets using a weight of 0.7 and 0.3, respectively. This partitioning ensures that the models are trained on sufficient data while retaining a separate portion for unbiased evaluation.

Target variable realisations are obtained from the data set \cite{kaggle}, which provides comprehensive football-related information, including scores, results, statistics, transfer news, and fixtures. These data were derived from the following resources available in the public domain: \url{https://www.kaggle.com/datasets/davidcariboo/player-scores}. As part of the feature engineering process, the features \emph{evaluation time of a player}, \emph{time lag between transfer value changes}, \emph{age of a player}, and \emph{age$^{2}$} are computed. Additionally, the \emph{pitch position of a player} is sourced from the \emph{StatsBomb} data set.

The outcomes of the trained models are presented in Table \ref{tab:transfer_value_predictors}, which showcases the performance and effectiveness of the different regression algorithms in predicting transfer values.
\begin{table}[h!]
    \tbl{Transfer value gain/loss predictors' results.}
    {\begin{tabular}{|P{4cm}|P{3cm}|P{2cm}|P{2cm}|}
    \hline
     & RMSE & MAE & Importance of \emph{player's score}(rank)\\
    \hline
    CatBoost Regressor & \textbf{6.23} & 4.53 & \textbf{1}\\
    \hline
    Random Forest Regressor & 6.29 & \textbf{4.26} & \textbf{1}\\
    \hline
    Decision Tree Regressor & 6.34 & 4.30 & 4\\
    \hline
    \end{tabular}}
    \label{tab:transfer_value_predictors}
\end{table}\par
It is essential to highlight that the range of the target variable in the test set differs by 37 (in millions of euros) between the minimum and maximum values. Due to large values within the test set, relying on the MAE metric is more appropriate as it is more robust to outliers compared to RMSE.

With this in mind, the \emph{Random Forest Regressor} algorithm performs better by achieving an MAE of 4.26 (in millions of euros). This indicates that, on average, the model’s predictions for gains or losses in transfer value deviate by 4.26 (in millions of euros).

When considering the absolute sum of the values in the test set, which amounts to
232.2 million euros, the proportion of the MAE result to the predicted values by the algorithm is merely 2.05\%. This demonstrates the high stability of the model, even when dealing with large value ranges.

The robustness and accuracy of the \emph{Random Forest Regressor} algorithm, as evidenced by its MAE performance, suggest its effectiveness in accurately predicting gains or losses in transfer values. 

\section{Results}
In this section, we present real-life applications of the resulting algorithm, which integrates the aforementioned models. Specifically, we focus on the \emph{European Football Championship 2020} and select three highly notable teams, namely \emph{Denmark}, \emph{Czech Republic} and \emph{Italy}. Subsequently, we randomly select two players from each team. 

For these selected footballers, their performance scores are computed and utilised as one of the parameters in the model established in Section \ref{Forecasting}. These performance scores provide valuable insights into the players’ on-the-field contributions, allowing us to incorporate their performance into the forecasting model. The results are presented in Table \ref{tab:transfer_value_changes}.\newpage
\begin{table}[h!]
    \tbl{Application of the algorithm.Transfer values changes.}
    {\begin{tabular}{|P{2cm}|P{2cm}|P{2cm}|P{2cm}|P{2cm}|P{2cm}|}
    \hline
     Name & Team & Position & Score & Predicted value change(mln. euros) & \emph{Kaggle data set} value change(mln. euros)\\
    \hline
    A. Belotti* & Italy & Center Forward & -2.08 & -3.08 & -7.00\\
    \hline
    Jorginho & Italy & Center Defensive Midfield & 1.51 & 15.30 & 5.00\\
    \hline
    M. Damsgaard & Denmark & Left Attacking Midfield & 1.17 & 11.20 & 7.00\\
    \hline
    J. Maehle & Denmark & Left Back & 2.05 & 7.20 & 6.00\\
    \hline
    O. Celustka* & Czech Republic & Right Center Back & 0.23 & -2.91 & -0.05\\
    \hline
    P. Schick & Czech Republic & Center Forward & 1.58 & 7.34 & 5.00\\
    \hline
    \end{tabular}}
    \begin{tablenotes}
        \small
        \item * - assessment considers only international performance due to data sparsity of national championships
    \end{tablenotes}
    \label{tab:transfer_value_changes}
\end{table}
By leveraging the performance scores of selected players, we can effectively evaluate the predictive capabilities of the established model in the context of the \emph{European Football Championship 2020}.\par
Furthermore, we apply our algorithm as a valuable tool for constructing the symbolic team of the \emph{European Football Championship 2020} with a 4-3-3 formation, consisting of 4 defenders, 3 midfielders, and 3 strikers. To accomplish this, we evaluated all teams that reached the quarter-final stage of the championship.

Based on their positions, we establish rankings for all players and select the top performers for each position. By considering their performances as determined by our algorithm, we assemble a highly skilled and representative team that encapsulates the tournament's standout players. This process ensures that the symbolic team comprises outstanding talents from various positions, forming a cohesive and formidable group in line with the 4-3-3 formation. 

As mentioned in Section \ref{method}, we leverage the normalised score, as defined in (\ref{eq:normalising}), to compare players. Additionally, we impose the condition that only footballers who had participated in three or more games throughout the entire championship are considered for evaluation.

By normalising the scores, we ensure a fair and equitable comparison among players, accounting for any variations or biases that might arise from game participation or playing time differences. This normalisation process allows us to assess players on a consistent scale and make meaningful comparisons based on their performance.

\begin{figure}[h!]
    \centering
    \includegraphics[scale=1]{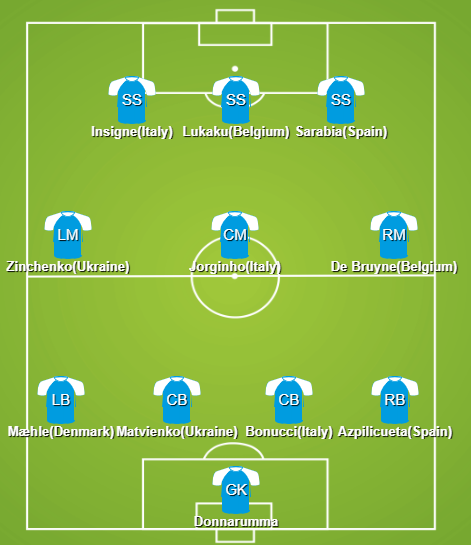}
    \caption{\centering{Team of the \emph{European Football Championship 2020}, formed by the algorithm\par
    (letters stand for the pitch position)}}
    \label{fig:symbolic_team}
\end{figure}

Furthermore, by setting the minimum threshold of three games played, we establish a criterion that guarantees a certain level of participation and consistency in performance. This condition offers a reliable basis for evaluating the players and ensures that our analysis encompasses players who have significantly contributed to their respective teams throughout the championship—the modelling results presented in Figure \ref{fig:symbolic_team}.
\newpage
These examples illustrate how the algorithm can be used to address specific challenges and generate insights in various domains. By applying the algorithm to real-life data sets, we showcase its ability to contribute to the understanding and analysis of complex systems.
\section{Discussion}
The current research can be enhanced from several perspectives. Firstly, incorporating optical-tracking data, which enables automatic tracking of all players' locations, would broaden the scope of actions that can be considered in the analysis. This richer data set would allow for a more comprehensive modelling of footballers’ actions, considering the actions of both the attacking and defending teams. Defensive actions could be evaluated by modelling the game as an \emph{action-reaction} system, enabling  a more holistic assessment of game states. This expanded model would encompass various scenarios and provide a more thorough understanding of the game dynamics.

Furthermore, the transfer value gain or loss prediction could be further improved by incorporating additional features into the model. For example, factors such as the remaining contract term, the economic level of the recruiting club \citep{Poli2022}, and the number of injuries for a player could enhance the predictive capabilities. By including these features, the model would gain insights into the influence of contract situations, club finances, and player fitness on transfer value dynamics.
Moreover, by collecting data from low- and middle-budget clubs, the research could achieve robustness and unbiasedness by accounting for a broader range of transfer fee patterns. Expanding the data set to include clubs of different economic capacities would provide valuable insights into the dynamics of transfer values across various contexts. It could uncover patterns that need to be more evident in data sets solely focused on high-budget clubs.


\section{Conclusions}
The results of this study showcase the models' potential for generating valuable insights into sports economics and PA. Managers can employ these models to optimise decision-making processes, particularly when managing budgets. Similarly, coaches and scouts can benefit from utilising the models as supplementary tools that offer data-driven perspectives on players. Therefore, applying the established models in the sports industry presents an avenue for practitioners to gain a competitive edge and make more informed and effective decisions.


\section*{Declaration of interest statement}
No potential conflict of interest was reported by the authors.

\section*{Funding details}
The authors reported there is no funding associated with the work featured in this article.


\newpage

\bibliographystyle{apalike}
\bibliography{interactapasample}

\begin{thebibliography}{20}
\providecommand{\natexlab}[1]{#1}
\providecommand{\url}[1]{\texttt{#1}}
\expandafter\ifx\csname urlstyle\endcsname\relax
  \providecommand{\doi}[1]{doi: #1}\else
  \providecommand{\doi}{doi: \begingroup \urlstyle{rm}\Url}\fi

\bibitem[Andersen et~al.(2022)Andersen, Francis, and Bateman]{Lasse2022}
L.~W. Andersen, J.~W. Francis, and M.~Bateman.
\newblock Danish association football coaches’ perception of performance
  analysis.
\newblock \emph{International Journal of Performance Analysis in Sport},
  22\penalty0 (1):\penalty0 149--173, 2022.
\newblock \doi{https://doi.org/10.1080/24748668.2021.2012040}.
\newblock URL \url{https://doi.org/10.1080/24748668.2021.2012040}.

\bibitem[Anzer and Bauer(2021)]{Anzer2021}
G.~Anzer and P.~Bauer.
\newblock A goal scoring probability model for shots based on synchronized
  positional and event data in football (soccer).
\newblock \emph{Frontiers in Sports and Active Living}, 3, 2021.
\newblock \doi{https://doi.org/10.3389/fspor.2021.624475}.
\newblock URL \url{https://doi.org/10.3389/fspor.2021.624475}.

\bibitem[Cronin et~al.(2012)Cronin, Bampouras, and Miller]{Miller2012}
C.~Cronin, T.~Bampouras, and P.~Miller.
\newblock Performance analytic processes in elite sport practice: An
  exploratory investigation of the perspectives of a sport scientist, coach and
  athlete.
\newblock \emph{International Journal of Performance Analysis in Sport},
  12:\penalty0 468--483, 2012.
\newblock \doi{https://doi.org/10.1080/24748668.2012.11868611}.
\newblock URL \url{https://doi.org/10.1080/24748668.2012.11868611}.

\bibitem[Decroos et~al.(2019)Decroos, Bransen, Van~Haaren, and Davis]{VAN}
T.~Decroos, L.~Bransen, J.~Van~Haaren, and J.~Davis.
\newblock Actions speak louder than goals: Valuing player actions in soccer.
\newblock In \emph{Proceedings of the 25th ACM SIGKDD International Conference
  on Knowledge Discovery \& Data Mining}, page 1851–1861, New York, NY, USA,
  2019. Association for Computing Machinery.
\newblock ISBN 9781450362016.
\newblock \doi{https://doi.org/10.1145/3292500.3330758}.
\newblock URL \url{https://doi.org/10.1145/3292500.3330758}.

\bibitem[Fernández et~al.(2019)Fernández, Bornn, and Cervon]{Javier2019}
J.~Fernández, L.~Bornn, and D.~Cervon.
\newblock Decomposing the immeasurable sport: A deep learning expected
  possession value framework for soccer.
\newblock In \emph{Research paper, MIT Sloan, Sport Analytics Conference,
  Boston}, 2019.

\bibitem[Goes-Smit et~al.(2018)Goes-Smit, Kempe, Meerhoff, and
  Lemmink]{Smit2018}
F.~Goes-Smit, M.~Kempe, R.~Meerhoff, and A.P.M.~Koen Lemmink.
\newblock Not every pass can be an assist: A data-driven model to measure pass
  effectiveness in professional soccer matches.
\newblock \emph{Big Data}, 7, 2018.
\newblock \doi{https://doi.org/10.1089/big.2018.0067}.
\newblock URL \url{https://doi.org/10.1089/big.2018.0067}.

\bibitem[He et~al.(2015)He, Cachucho, and Knobbe]{Miao}
M.~He, R.~Cachucho, and A.~J. Knobbe.
\newblock Football player's performance and market value.
\newblock In \emph{In Proceedings of the 2nd workshop of sports analytics,
  European Conference on Machine Learning and Principles and Practice of
  Knowledge Discovery in Databases (ECML PKDD)}, pages 87--95, 2015.

\bibitem[Kaggle(2023)]{kaggle}
Kaggle.
\newblock Football data from transfermarkt, 2023.
\newblock URL \url{https://www.kaggle.com/datasets/davidcariboo/player-scores}.

\bibitem[Kovacs and Toka(2022)]{Roland2022}
R.~Kovacs and L.~Toka.
\newblock Predicting player transfers in the small world of football, 2022.
\newblock URL \url{https://doi.org/10.1007/978-3-031-02044-5\_4}.

\bibitem[Liu and Schulte(2018)]{Guiliang2018}
G.~Liu and O.~Schulte.
\newblock Deep reinforcement learning in ice hockey for context-aware player
  evaluation.
\newblock In \emph{Proceedings of the Twenty-Seventh International Joint
  Conference on Artificial Intelligence, July, IJCAI-18.}, pages 3442--3448,
  2018.
\newblock \doi{https://doi.org/10.24963/ijcai.2018/478}.
\newblock URL \url{https://doi.org/10.24963/ijcai.2018/478}.

\bibitem[Liu et~al.(2020)Liu, Luo, Schulte, and Kharrat]{Liu2020}
G.~Liu, Y.~Luo, O.~Schulte, and T.~Kharrat.
\newblock Deep soccer analytics: Learning an action-value function for
  evaluating soccer players.
\newblock \emph{Data Min. Knowl. Discov.}, 34\penalty0 (5):\penalty0
  1531–1559, 2020.
\newblock ISSN 1384-5810.
\newblock \doi{https://doi.org/10.1007/s10618-020-00705-9}.
\newblock URL \url{https://doi.org/10.1007/s10618-020-00705-9}.

\bibitem[Lucey et~al.(2015)Lucey, Bialkowski, Monfort, Carr, and
  Matthews]{Lucey2015QualityVQ}
P.~Lucey, A.~Bialkowski, M.~Monfort, P.~Carr, and I.~Matthews.
\newblock "{Q}uality vs quantity": Improved shot prediction in soccer using
  strategic features from spatiotemporal data.
\newblock In \emph{In Proceedings of 8th Annual MIT Sloan Sports Analytics
  Conference (pp. 1-9)}, 2015.

\bibitem[McHale and Holmes(2023)]{Ian2023}
I.~G. McHale and B.~Holmes.
\newblock Estimating transfer fees of professional footballers using advanced
  performance metrics and machine learning.
\newblock \emph{European Journal of Operational Research}, 306\penalty0
  (1):\penalty0 389--399, 2023.
\newblock ISSN 0377-2217.
\newblock \doi{https://doi.org/10.1016/j.ejor.2022.06.033}.
\newblock URL \url{https://doi.org/10.1016/j.ejor.2022.06.033}.

\bibitem[Nicholls et~al.(2018)Nicholls, James, Bryant, and Wells]{Nicholls2018}
S.~Nicholls, N.~James, E.~Bryant, and J.~Wells.
\newblock Elite coaches’ use and engagement with performance analysis within
  olympic and paralympic sport.
\newblock \emph{International Journal of Performance Analysis in Sport},
  18:\penalty0 1--16, 2018.
\newblock \doi{https://doi.org/10.1080/24748668.2018.1517290}.
\newblock URL \url{https://doi.org/10.1080/24748668.2018.1517290}.

\bibitem[Patnaik et~al.(2019)Patnaik, Praharaj, Prakash, and
  Samdani]{Patnaik2019}
D.~Patnaik, H.~Praharaj, K.~Prakash, and K.~Samdani.
\newblock A study of prediction models for football player valuations by
  quantifying statistical and economic attributes for the global transfer
  market.
\newblock In \emph{2019 IEEE International Conference on System, Computation,
  Automation and Networking (ICSCAN)}, pages 1--7, 2019.
\newblock \doi{https://doi.org/10.1109/ICSCAN.2019.8878843}.
\newblock URL \url{https://doi.org/10.1109/ICSCAN.2019.8878843}.

\bibitem[Poli et~al.(2022)Poli, Besson, and Ravenel]{Poli2022}
R.~Poli, R.~Besson, and L.~Ravenel.
\newblock Econometric approach to assessing the transfer fees and values of
  professional football players.
\newblock \emph{Economies}, 10\penalty0 (1), 2022.
\newblock ISSN 2227-7099.
\newblock \doi{https://doi.org/10.3390/economies10010004}.
\newblock URL \url{https://doi.org/10.3390/economies10010004}.

\bibitem[Power et~al.(2017)Power, Ruiz, Wei, and Lucey]{Power2017}
P.~Power, H.~Ruiz, X.~Wei, and P.~Lucey.
\newblock Not all passes are created equal: Objectively measuring the risk and
  reward of passes in soccer from tracking data.
\newblock In \emph{Proceedings of the 23rd ACM SIGKDD International Conference
  on Knowledge Discovery and Data Mining}, pages 1605--1613, 2017.
\newblock ISBN 978-1-4503-4887-4.
\newblock \doi{https://doi.org/10.1145/3097983.3098051}.
\newblock URL \url{https://doi.org/10.1145/3097983.3098051}.

\bibitem[Rahimian et~al.(2022)Rahimian, Van~Haaren, Abzhanova, and
  Toka]{Rahimian2022}
P.~Rahimian, J.~Van~Haaren, T.~Abzhanova, and L.~Toka.
\newblock Beyond action valuation: A deep reinforcement learning framework for
  optimizing player decisions in soccer.
\newblock In \emph{16th Annual MIT Sloan Sports Analytics Conference. Boston,
  MA, USA: MIT}, page~25, 2022.

\bibitem[Rein et~al.(2017)Rein, Raabe, and Memmert]{REIN2017}
R.~Rein, D.~Raabe, and D.~Memmert.
\newblock “{W}hich pass is better ?” novel approaches to assess passing
  effectiveness in elite soccer.
\newblock \emph{Human Movement Science}, 55:\penalty0 172--181, 2017.
\newblock ISSN 0167-9457.
\newblock \doi{https://doi.org/10.1016/j.humov.2017.07.010}.
\newblock URL \url{https://doi.org/10.1016/j.humov.2017.07.010}.

\bibitem[StatsBomb(2023)]{statsbomb}
StatsBomb.
\newblock open-data, 2023.
\newblock URL \url{https://github.com/statsbomb/open-data}.

\end{thebibliography}

\end{document}